\documentclass{article}

\usepackage{microtype}
\usepackage{graphicx}
\usepackage{subcaption}
\usepackage{booktabs} 

\usepackage{hyperref}

\usepackage[preprint]{icml2026}

\usepackage{amsmath}
\usepackage{amssymb}
\usepackage{mathtools}
\usepackage{amsthm}

\usepackage{algorithm}
\usepackage{algorithmic}
\usepackage[capitalize,noabbrev]{cleveref}

\theoremstyle{plain}

\theoremstyle{definition}

\theoremstyle{remark}

\usepackage[disable,textsize=tiny]{todonotes}

\icmltitlerunning{Reasoning New Skills from Existing Abilities for Cross-Task Robotic Manipulation}

\begin{document}

\twocolumn[
  \icmltitle{Decompose and Recompose: Reasoning New Skills from Existing Abilities for Cross-Task Robotic Manipulation}



  \icmlsetsymbol{equal}{*}
    \begin{icmlauthorlist}
    \icmlauthor{Xitie Zhang}{tju}
    \icmlauthor{Aming Wu}{hfut}
    \icmlauthor{Yahong Han}{tju}
    \end{icmlauthorlist}
    
    \icmlaffiliation{tju}{School of Artificial Intelligence, College of Intelligence and Computing, Tianjin University, China}
    \icmlaffiliation{hfut}{School of Computer Science and Information Engineering, Hefei University of Technology, China}

    
    \icmlcorrespondingauthor{Yahong Han}{yahong@tju.edu.cn}





  \vskip 0.3in
]



\printAffiliationsAndNotice{}  

\begin{abstract}
Cross-task generalization is a core challenge in open-world robotic manipulation, 
and the key lies in extracting transferable manipulation knowledge from seen tasks. 
Recent in-context learning approaches leverage seen task demonstrations to generate 
actions for unseen tasks without parameter updates. However, existing methods provide 
only low-level continuous action sequences as context, failing to capture composable 
skill knowledge and causing models to degenerate into superficial trajectory imitation. We propose \emph{Decompose and Recompose}, a skill reasoning 
framework using atomic skill--action pairs as intermediate 
representations. Our approach decomposes seen demonstrations 
into interpretable skill--action alignments, enabling the 
model to recompose these skills for unseen tasks through 
compositional reasoning.  Specifically, we construct a task-adaptive dynamic demonstration library via visual-semantic retrieval 
combined with skill sequences from a planning agent, complemented by a coverage-aware 
static library to fill missing skill patterns. 
Together, these yield skill-comprehensive demonstrations that explicitly elicit 
compositional reasoning for skill composition and execution ordering. Experiments 
on the AGNOSTOS benchmark and real-world environments validate our method's zero-shot 
cross-task generalization capability.
\end{abstract}

\section{Introduction}

\begin{figure*}[!t]
  \begin{center}
    \centerline{\includegraphics[width=\textwidth]{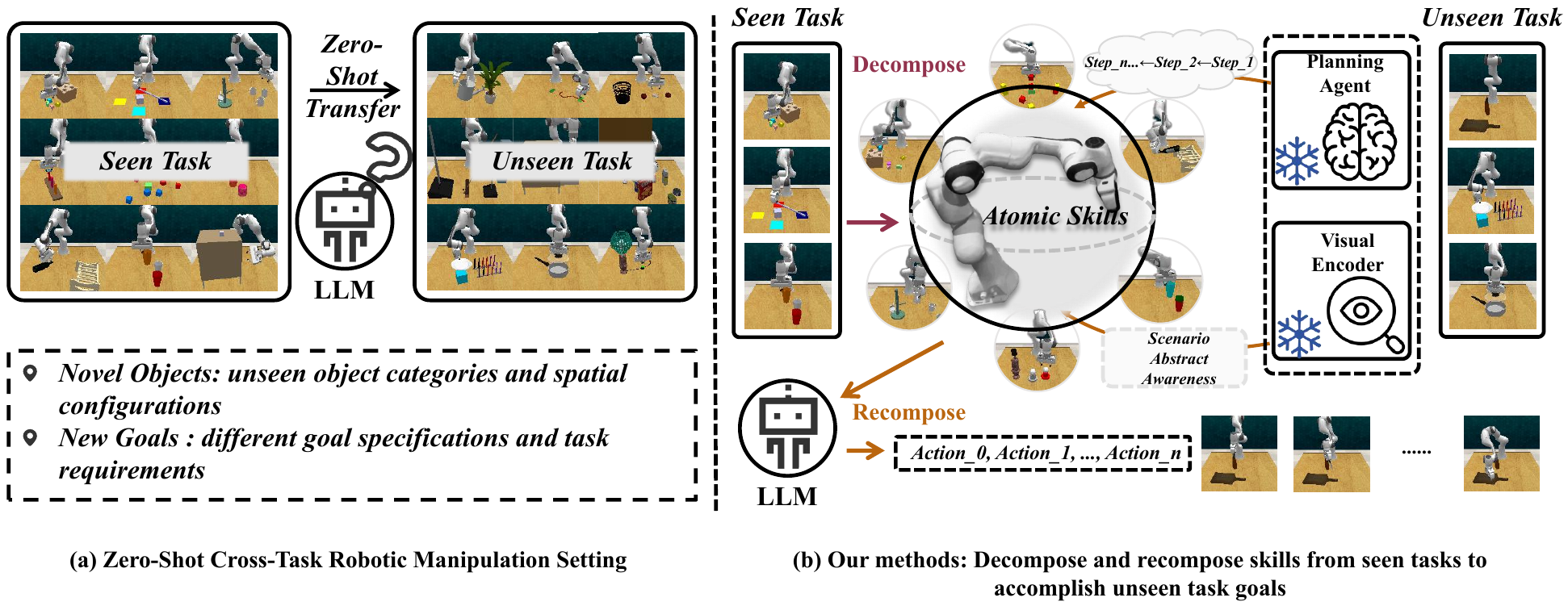}}
    \caption{
    Overview of cross-task robotic manipulation. (a) The zero-shot cross-task setting requires transferring knowledge from seen tasks to unseen tasks involving novel objects and new goals. (b) Our \emph{Decompose and Recompose} framework extracts atomic skills from seen task demonstrations, uses a planning agent to predict skill sequences for unseen tasks, and leverages visual encoding for scene-aware retrieval. The LLM then reasons over skill--action aligned demonstrations to generate executable action sequences for the target task.
    }
    \label{head}
  \end{center}
\vspace{-0.6cm}
\end{figure*}

With the rapid development of Vision-Language-Action (VLA) 
models~\cite{brohan2023rt,kim2024openvla,black2024pi_0,team2024octo}, 
robotic manipulation has achieved significant progress in handling 
visual perturbations within known tasks. However, real-world 
deployment inevitably involves novel objects, new goals, and 
action compositions that were never observed. 
\cite{zhou2025exploring} introduces 
the cross-task generalization setting, which requires robots 
to transfer manipulation knowledge from seen tasks to entirely 
unseen tasks without parameter updates, as illustrated in 
Figure~\ref{head}(a). Compared to within-task generalization 
that focuses on visual robustness under fixed task semantics, 
cross-task generalization requires transferable skill understanding, 
which is essential for deploying general-purpose robots in 
open-world environments.

The core challenge of cross-task generalization lies in extracting 
transferable manipulation knowledge from seen tasks and reasoning 
over unseen tasks. Existing methods such as X-ICM~\cite{zhou2025exploring} 
train a diffusion-based dynamics-guided sample selection module to 
retrieve demonstrations from seen tasks based on dynamics similarity, 
and use them to prompt LLMs for action prediction. However, these 
methods require training on specific task distributions, weakening 
cross-domain transferability, and provide only low-level numerical 
action sequences as context, relying on direct alignment of trajectory 
shapes without explicit skill semantics. Such numerical representations 
fail to express causal and procedural information, including what 
skill is being performed at each step, why it is executed, and how 
it connects to subsequent steps. This causes LLMs to only perform 
trajectory pattern matching rather than skill reasoning, limiting 
their generalization capability to tasks with novel skill compositions.

To address these issues, as shown in Figure~\ref{head}(b), 
we propose \emph{Decompose and Recompose}, a compositional 
skill reasoning framework that reasons new skills from 
existing abilities for cross-task manipulation generalization. 
We decompose each task demonstration into atomic skill 
sequences and atomic skill label--action alignment pairs, 
where the former enables skill sequence similarity retrieval 
and reasoning, and the latter provides composable skill 
structures for inference, while maintaining executable 
low-level actions as the final output. This elevates 
cross-task transfer from trajectory shape similarity to 
composable skill structures, explicitly activating the 
model's reasoning capabilities for skill composition and 
execution ordering.

Specifically, we employ a general vision model~\cite{simeoni2025dinov3} to extract static visual-semantic representations for scene similarity retrieval, and use a planning agent to predict atomic skill sequences for unseen tasks. Based on these predicted sequences, we retrieve and rerank demonstrations with similar action logic from the candidate pool, constructing a task-adaptive dynamic demonstration library. However, relying solely on dynamic retrieval may still result in incomplete demonstration coverage: even if the retrieved demonstrations have high overall similarity, they may lack critical skill patterns required for task completion. To address this, we introduce an offline-constructed coverage-aware static demonstration library, converting planner-output skill sequences into structured tokens (e.g., verb skills, skill chains) and normalizing the skill token space through Inverse Document Frequency (IDF) based Token Weighting and Length Regularization strategies. During inference, we compute the covered token set from the dynamic library, explicitly identify gaps, and greedily supplement missing tokens from the static library. This yields demonstrations that are both task-adaptive and skill-coverage sufficient, making in-context demonstrations a more helpful skill evidence collection for unseen tasks, thereby enhancing cross-task compositional generalization.

In summary, our main contributions are as follows:
\begin{itemize}
    \item We construct atomic skill label-action aligned data, providing LLMs with composable and interpretable atomic skill sets that enable structured reasoning over manipulation sequences.
   \item We propose \emph{Decompose and Recompose}, a compositional skill reasoning framework for zero-shot cross-task manipulation that obtains task-adaptive and skill-comprehensive demonstration sets through the synergy of a dynamic demonstration library for task relevance and a static demonstration library for skill coverage.
    \item We validate our method's effectiveness on cross-task zero-shot generalization toward unseen tasks on the AGNOSTOS benchmark and real-world environments, achieving competitive performance.
\end{itemize}

\begin{figure*}[ht!]
  \begin{center}
    \centerline{\includegraphics[width=\textwidth]{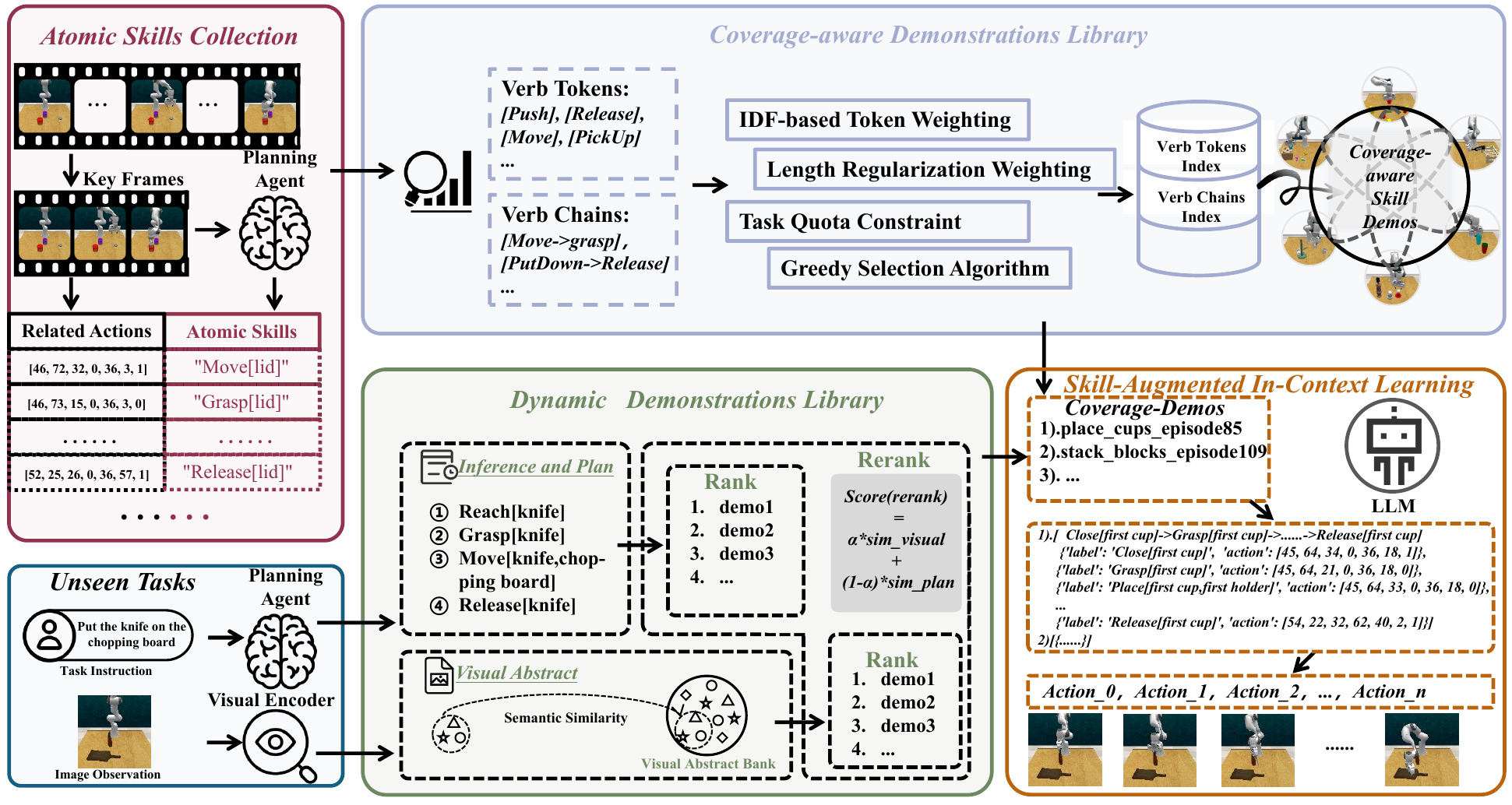}}
    \caption{
 Overview of our method. Our framework consists of four components: (1) \textbf{Atomic Skills Collection} extracts skill--action pairs from seen demonstrations as composable intermediate representations; (2) \textbf{Coverage-aware Static Library} uses IDF-based token weighting to ensure skill pattern coverage; (3) \textbf{Dynamic Demonstrations Library} retrieves task-adaptive examples via visual and plan-based similarity; (4) \textbf{Skill-Augmented In-Context Learning} feeds retrieved demonstrations to an LLM for compositional skill reasoning to predict action sequences.
    }
    \label{main}
  \end{center}
\vspace{-0.4cm}
\end{figure*}

\section{Related Works}

\textbf{Vision-Language-Action Models.} Vision-Language-Action (VLA) models enable robots to understand instructions and interact with the physical world, achieving significant progress in general-purpose manipulation. Existing VLA models primarily follow two paradigms. Modular approaches decompose perception, language understanding, planning, and action execution into separate components, such as VoxPoser~\cite{huang2023voxposer}, MOKA~\cite{liu2024moka}, COPA~\cite{huang2024copa}, and ReKep~\cite{huang2024rekep}, which typically require task-specific prompt engineering or hand-crafted designs. End-to-end approaches train policy models to directly map raw sensory inputs to robot actions, with success heavily dependent on training data scale and diversity. Representative works include RT-2~\cite{brohan2023rt} and RT-2-X~\cite{vuong2023open} trained on large-scale robotic datasets, OpenVLA~\cite{kim2024openvla} as the first fully open-source VLA model, and recent advances like $\pi_0$~\cite{black2024pi_0}, RDT~\cite{liu2024rdt}, LLARVA~\cite{niu2024llarva}, and HPT~\cite{wang2024hpt} that incorporate sophisticated architectures and training objectives. Despite progress in visual robustness within known tasks, existing VLA models remain limited in zero-shot cross-task generalization. AGNOSTOS~\cite{zhou2025exploring} provides the first systematic evaluation, revealing widespread degradation on unseen tasks and indicating that large-scale pre-training alone is insufficient for cross-task transfer.

\textbf{In-Context Learning.} Large language models demonstrate remarkable in-context learning (ICL) capabilities~\cite{brown2020language}, generalizing to new tasks from few demonstrations without parameter updates. ICL has been successfully applied across various domains~\cite{dong2022survey,wies2023learnability,rubin2021learning,min2022rethinking,zhang2023makes}. In robotics, RoboPrompt~\cite{yin2024context} first showed that text-only LLMs can directly predict robot actions through ICL by converting episodes into textual descriptions. Subsequent works like KAT~\cite{di2024keypoint}, InCoRo~\cite{zhu2024incoro}, and Instant Policy~\cite{vosylius2024instant} further explored ICL for robotic control, but primarily focused on within-task generalization. Since demonstration quality significantly affects ICL performance~\cite{rubin2021learning,min2022rethinking}, and long contexts may cause degradation~\cite{dong2022survey,wies2023learnability}, selecting effective examples is crucial. Recently, X-ICM~\cite{zhou2025exploring} first extended ICL to the zero-shot cross-task setting with dynamics-guided sample selection. However, it provides only low-level action sequences as context, which lack explicit composable skill structures and causal information, causing LLMs to degrade into surface-similarity-based trajectory imitation. In contrast, our method constructs aligned representations of atomic skill labels and low-level actions, providing composable and interpretable intermediate structures that elevate cross-task transfer from "trajectory similarity" to "skill composability," explicitly activating reasoning over skill composition and execution order.

\section{Method}

We propose a framework for zero-shot cross-task robotic manipulation that addresses the fundamental challenge of transferring operational knowledge from seen tasks to unseen tasks. Unlike prior approaches that rely on training dynamics-guided retrievers on specific task distributions, our method achieves cross-task generalization through a training-free pipeline that leverages pretrained visual encoders and large language models. The core insight is that providing LLMs with composable and interpretable intermediate representations---atomic skill labels aligned with low-level actions---is more effective for cross-task transfer than simply presenting raw numerical action sequences.

As illustrated in Figure~\ref{main}, our approach consists of three tightly integrated components: (1) \textbf{atomic skill collection} that constructs label-action aligned data for seen task demonstrations; (2) \textbf{dual-library demonstration retrieval} that combines visual scene similarity with plan-based skill coverage; and (3) \textbf{skill-augmented in-context learning} that enables compositional reasoning while outputting executable low-level actions.

\subsection{Problem Formulation}

Consider a robot manipulation setting where we have access to a demonstration library $\mathcal{D}^{\text{seen}}$ collected from a set of seen tasks $\mathcal{T}^{\text{seen}}$. Each demonstration $d \in \mathcal{D}^{\text{seen}}$ consists of a task instruction $\ell_d$, a sequence of keyframe observations $(o_0^d, o_1^d, \ldots, o_K^d)$ extracted based on gripper state changes and motion stopping criteria, and the corresponding 7-DoF actions $(a_1^d, \ldots, a_K^d)$. Each action $a_t$ specifies discretized end-effector position (3D voxel indices), rotation (Euler angles discretized into bins), and binary gripper state.

Given an unseen task $\tau^{\text{unseen}} \notin \mathcal{T}^{\text{seen}}$ with instruction $\ell^q$ and initial observation $o_0^q$, our goal is to generate an action sequence $\{a_1^q, \ldots, a_T^q\}$ that successfully accomplishes the task. The key challenge lies in the distribution shift: unseen tasks may involve novel object configurations, different goal specifications, or new combinations of primitive skills not directly observed during training.

\subsection{Atomic Skill Collection}

\begin{figure}[!t]
  \begin{center}
    \centerline{\includegraphics[width=\columnwidth]{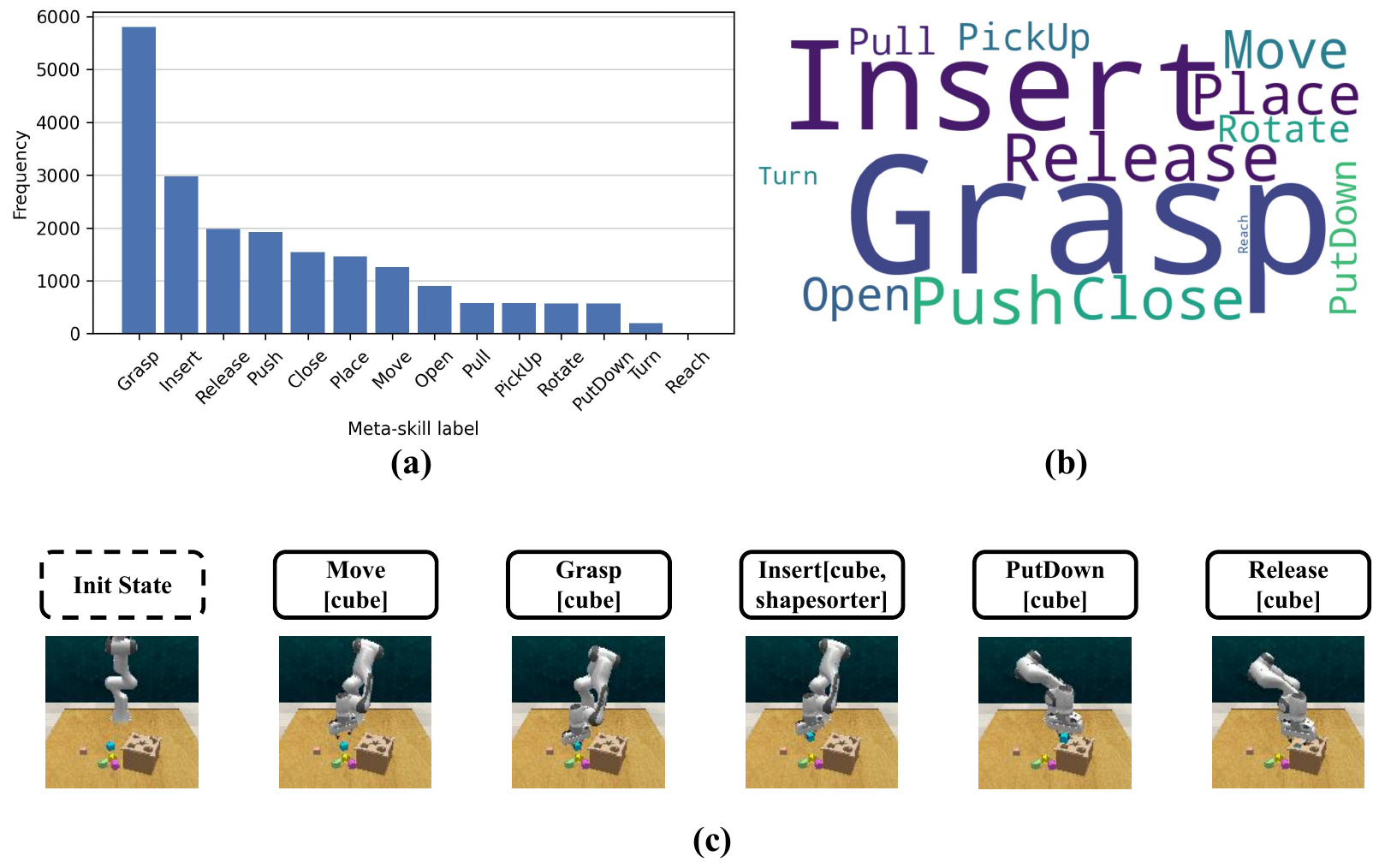}}
    \caption{
 Atomic Skill Collection. (a) Frequency distribution of extracted atomic skills across seen demonstrations. (b) Word cloud of the atomic skill vocabulary. (c) Visualization of keyframe detection and atomic skill labeling for a sample demonstration, showing the progression from initial state through intermediate keyframes with corresponding skill annotations.
    }
    \label{skillcollection}
  \end{center}
   \vspace{-0.6cm}
\end{figure}

To enable compositional reasoning, we construct atomic skill labels for each demonstration in the seen task library, transforming opaque numerical action sequences into interpretable skill--action pairs. As illustrated in Figure~\ref{skillcollection}, our pipeline extracts keyframes from demonstrations, annotates each segment with atomic skill labels, and produces a diverse vocabulary of reusable skills.

\textbf{Keyframe Extraction.} Given a demonstration episode, we identify keyframes using: (1) gripper state changes (open to closed or vice versa); (2) motion stopping events, detected when joint velocities fall below a threshold; and (3) episode termination. This produces segmented demonstrations where each segment $(o_k, o_{k+1}, a_k)$ represents a coherent atomic action.

\textbf{VLM-based Annotation.} For each action segment, we employ a vision-language model to annotate it with an atomic skill label following the format $\texttt{Verb}[\texttt{obj}]$ or $\texttt{Verb}[\texttt{obj}_1, \texttt{obj}_2]$, where:
\begin{equation}
\texttt{Verb} \in \mathcal{V} = \{\texttt{Reach}, \texttt{Move}, \texttt{Grasp}, \texttt{Release}, \ldots\}
\end{equation}
The VLM receives the start and end keyframe images $(o_k, o_{k+1})$, gripper state transition $(g_k, g_{k+1})$, object names from segmentation masks, and task instruction.

\textbf{Gripper-Constrained Labeling.} We leverage gripper transitions as hard constraints: when the gripper changes from open to closed ($g_k=1, g_{k+1}=0$), the label is forced to be $\texttt{Grasp}[\cdot]$; when it changes from closed to open, the label is $\texttt{Release}[\cdot]$. This ensures physical consistency and reduces annotation errors.

\textbf{Post-processing.} We apply rule-based post-processing: (1) for relational verbs like $\texttt{Close}$, $\texttt{Insert}$, $\texttt{Place}$, we enforce argument order as (movable\_object, target\_object); (2) for $\texttt{Grasp}$/$\texttt{Release}$, we verify the object belongs to the movable category; (3) when the gripper remains open during a relational action, we downgrade to $\texttt{Move}$. This produces clean label-action pairs $\{(s_k, a_k)\}_{k=1}^K$.

\subsection{Dual-Library Demonstration Retrieval}

We address the challenge of selecting demonstrations that are both relevant and skill-complete through a dual-library system: a dynamic library for task-adaptive retrieval and a static library for skill coverage completion.

\subsubsection{Dynamic Demonstrations Library}

The dynamic library retrieves task-relevant demonstrations by fusing visual similarity with plan-based skill matching. For visual similarity, we extract a global feature from the query observation $o^q$ using a pretrained DINOv3~\cite{simeoni2025dinov3} encoder: $\mathbf{f}^q = \phi_{\text{vis}}(o^q)/\|\phi_{\text{vis}}(o^q)\|_2$, and compute cosine similarity $s_i^{\text{vis}} = \mathbf{f}^q \cdot \mathbf{f}_i$ with each demonstration $d_i$.

For plan-based similarity, a planning agent $\pi_{\text{plan}}$ first generates a predicted atomic skill sequence $\hat{\mathcal{S}} = (\hat{s}_1, \ldots, \hat{s}_M)$ based on the instruction and scene state, e.g., discretized object coordinates and gripper state, and more details in Appendix~\ref{canshudingyi}. We then measure the alignment between the predicted plan and each demonstration's skill labels $\mathcal{S}_i$ using Jaccard similarity over both verb sets and bigram chains:
\begin{equation}
s_i^{\text{plan}} = \lambda \cdot J(\mathcal{V}(\hat{\mathcal{S}}), \mathcal{V}(\mathcal{S}_i)) + (1-\lambda) \cdot J(\mathcal{B}(\hat{\mathcal{S}}), \mathcal{B}(\mathcal{S}_i))
\end{equation}
where $\mathcal{V}(\cdot)$ extracts verb sets and $\mathcal{B}(\cdot)$ extracts verb bigrams. The final ranking score fuses both similarities as $s_i = \alpha \cdot \tilde{s}_i^{\text{vis}} + (1-\alpha) \cdot s_i^{\text{plan}}$, where $\tilde{s}_i^{\text{vis}}$ is min-max normalized. We select the top-$k_{\text{sim}}$ demonstrations to form $\mathcal{D}_{\text{dyn}}$.

\subsubsection{Coverage-Aware Demonstrations Library}

While the dynamic library provides task-relevant demonstrations, it may still lack critical skill patterns required for task completion. To ensure comprehensive skill coverage, we maintain an offline-constructed Coverage-aware Demonstrations Library $\mathcal{L}_{\text{cov}}$ that supplements missing skill patterns through object-agnostic coverage tokens and IDF-weighted selection.

Each demonstration is represented by tokens that abstract away object identities, capturing only the verb-level skill structure:
\begin{equation}
\mathcal{T}(d) = \{\texttt{V:}v \mid v \in \mathcal{V}(\mathcal{S}_d)\} \cup \{\texttt{B:}v_1\!\rightarrow\!v_2 \mid (v_1,v_2) \in \mathcal{B}(\mathcal{S}_d)\}
\end{equation}
To prioritize rare but potentially critical skills, we assign IDF-based weights to each token: $w_t = (\log\frac{N+1}{\text{df}(t)+1} + 1)^\beta$, where $N$ is the total number of demonstrations, $\text{df}(t)$ is document frequency, and $\beta$ controls the weighting intensity. The selection score balances coverage gain against demonstration length:
\begin{equation}
\text{Score}(d) = \frac{\sum_{t \in \mathcal{T}(d) \setminus \mathcal{C}} w_t}{1 + \gamma \cdot |\mathcal{S}_d|}
\end{equation}
where $\mathcal{C}$ denotes currently covered tokens and $\gamma$ is a length penalty.

During inference, we compute the coverage gap $\mathcal{G} = \mathcal{T}(\hat{\mathcal{S}}) \setminus \bigcup_{d \in \mathcal{D}_{\text{dyn}}} \mathcal{T}(d)$ between the predicted plan's required tokens and those already covered by the dynamic library. We then greedily select up to $k_{\text{cov}}$ demonstrations from $\mathcal{L}_{\text{cov}}$ to fill missing tokens, yielding the final demonstration set $\mathcal{D} = \mathcal{D}_{\text{dyn}} \cup \mathcal{D}_{\text{cov}}$.

\begin{table*}[!ht]
\caption{Cross-task zero-shot manipulation performance on 23 unseen tasks from the AGNOSTOS benchmark. We report success rates (\%) for each task, grouped by difficulty level (Level-1 and Level-2). Column headers show task abbreviations (full names in Appendix). N/A indicates tasks overlapping with training data. Best results are in \textbf{bold}, second best marked with *.}
\label{tab:main_results}
\begin{center}
\begin{small}
\setlength{\tabcolsep}{2.5pt}
\resizebox{\textwidth}{!}{
\begin{tabular}{ll|ccccccccccccc}
\toprule
& & \multicolumn{13}{c}{\textbf{Level-1 Tasks}} \\
\cmidrule(lr){3-15}
\textbf{Category} & \textbf{Method} & \textit{Toilet} & \textit{Knife} & \textit{Fridge} & \textit{Micro.} & \textit{Laptop} & \textit{Phone} & \textit{Seat} & \textit{LampOff} & \textit{LampOn} & \textit{Book} & \textit{Umb.} & \textit{Grill} & \textit{Bin} \\
\midrule
& PerAct & 0.0 & 5.3 & 37.3 & 64.0 & 2.7 & 0.0 & 72.0 & 0.0 & 1.3 & 0.0 & 1.3 & 8.0 & 54.7 \\
& RVT & 0.0 & 2.7 & 50.7 & 26.7 & 50.7 & 2.7 & 40.0 & 0.0 & 1.3 & 0.0 & 1.3 & 0.0 & 6.7 \\
In-Domain & Sigma-Agent & 0.0 & 9.3 & 56.0 & 9.3 & 30.7 & 1.3 & 65.3 & 1.3 & 0.0 & 0.0 & 0.0 & 1.3 & 4.0 \\
& RVT2 & 0.0 & 1.3 & 0.0 & 17.3 & 42.7 & 1.3 & 62.7 & 2.7 & 1.3 & 0.0 & 1.3 & 5.3 & 34.7 \\
& InstantPolicy & 0.0 & 1.3 & 13.3 & 4.0 & 4.0 & 18.7 & 24.0 & 0.0 & 0.0 & 0.0 & 0.0 & 0.0 & 0.0 \\
\midrule
& D4R & 0.0 & 8.0 & 32.0 & 30.7 & 24.0 & 0.0 & 65.3 & 20.0 & 4.0 & 0.0 & 0.0 & 0.0 & 0.0 \\
Human-Video & R3M & 0.0 & 0.0 & 37.3 & 22.7 & 25.3 & 1.3 & 62.7 & 6.7 & 4.0 & 0.0 & 0.0 & 0.0 & 0.0 \\
& D4R-Align & 0.0 & 2.7 & 45.3 & 74.7 & 24.0 & 0.0 & 41.3 & 0.0 & 0.0 & 1.3 & 0.0 & 0.0 & 0.0 \\
& R3M-Align & 0.0 & 4.0 & 49.3 & 25.3 & 21.3 & 0.0 & 49.3 & 0.0 & 5.3 & 0.0 & 0.0 & 1.3 & 1.3 \\
\midrule
& OpenVLA & 0.0 & 5.3 & 38.7 & 40.0 & 57.3 & 0.0 & 53.3 & 12.0 & 1.3 & 1.3 & 0.0 & 10.7 & 0.0 \\
& RDT & 0.0 & 0.0 & 46.7 & 13.3 & 14.7 & 0.0 & 50.7 & 0.0 & 0.0 & 1.3 & 0.0 & 8.0 & 0.0 \\
& $\pi_0$ & 0.0 & 5.3 & 85.3 & 24.0 & 40.0 & 1.3 & 64.0 & 18.7 & 8.0 & 1.3 & 0.0 & 33.3 & 1.3 \\
Foundation & LLARVA & 0.0 & 0.0 & 12.0 & 0.0 & 6.7 & 0.0 & 40.0 & 0.0 & 0.0 & 0.0 & 0.0 & 0.0 & 0.0 \\
VLA & 3D-LOTUS & 0.0 & 6.7 & N/A & N/A & N/A & 0.0 & 6.7 & 0.0 & 0.0 & 0.0 & 0.0 & 13.3 & 5.3 \\
& 3D-LOTUS++ & 0.0 & 5.3 & N/A & N/A & N/A & 9.3 & 68.0 & 10.7 & 0.0 & 0.0 & 0.0 & 29.3 & 13.3 \\
& SAM2Act & 0.0 & 0.0 & 36.0 & 40.0 & 6.7 & 6.7 & 62.7 & 6.7 & 0.0 & 1.3 & 1.3 & 9.3 & 0.0 \\
& VoxPoser & 0.0 & 0.0 & 0.0 & 0.0 & 5.3 & 8.0 & 28.0 & 88.7 & 25.3 & 0.0 & 0.0 & 0.0 & 82.7 \\
\midrule
ICL-based & X-ICM  & 1.3 & 26.7 & 22.7 & 45.3 & 33.3 & 57.3 & 48.0 & 58.7 & 50.7 & 1.3 & 0.0 & 8.0 & 18.7 \\
\midrule
 & \textbf{Ours } & 1.3 & 21.3 & 34.7 & 62.7 & 34.7 & 42.7 & 72.0 & 67.0 & 52.3 & 1.3 & 0.0 & 12.0 & 20.0 \\
\bottomrule
\end{tabular}
}
\end{small}
\end{center}

\begin{center}
\begin{small}
\setlength{\tabcolsep}{2.5pt}
\resizebox{\textwidth}{!}{
\begin{tabular}{ll|cccccccccc|ccc}
\toprule
& & \multicolumn{10}{c|}{\textbf{Level-2 Tasks}} & \multicolumn{3}{c}{\textbf{Average}} \\
\cmidrule(lr){3-12} \cmidrule(lr){13-15}
\textbf{Category} & \textbf{Method} & \textit{USB} & \textit{Lid} & \textit{Plate} & \textit{Ball} & \textit{Scoop} & \textit{Rope} & \textit{Oven} & \textit{Buzz} & \textit{Plants} & \textit{Charger} & \textbf{L-1} & \textbf{L-2} & \textbf{All} \\
\midrule
& PerAct & 58.7 & 2.7 & 0.0 & 0.0 & 0.0 & 0.0 & 1.3 & 4.0 & 6.7 & 2.7 & 19.0$_{\pm 1.4}$ & 7.6$_{\pm 1.1}$ & 14.0$_{\pm 0.9}$ \\
& RVT & 89.3 & 2.7 & 0.0 & 0.0 & 0.0 & 0.0 & 4.0 & 8.0 & 5.3 & 4.0 & 14.0$_{\pm 1.4}$ & 11.3$_{\pm 1.6}$ & 12.8$_{\pm 0.2}$ \\
In-Domain & Sigma-Agent & 88.0 & 0.0 & 0.0 & 0.0 & 0.0 & 0.0 & 4.0 & 8.0 & 5.3 & 1.3 & 13.7$_{\pm 1.6}$ & 10.7$_{\pm 1.7}$ & 12.4$_{\pm 0.4}$ \\
& RVT2 & 22.7 & 40.0 & 0.0 & 0.0 & 0.0 & 0.0 & 0.0 & 1.3 & 1.3 & 1.3 & 13.1$_{\pm 0.4}$ & 6.7$_{\pm 1.3}$ & 10.3$_{\pm 0.6}$ \\
& InstantPolicy & 26.7 & 1.3 & 0.0 & 0.0 & 0.0 & 0.0 & 0.0 & 1.3 & 0.0 & 0.0 & 4.3$_{\pm 4.2}$ & 2.9$_{\pm 1.4}$ & 3.7$_{\pm 3.0}$ \\
\midrule
& D4R & 98.7 & 0.0 & 0.0 & 0.0 & 0.0 & 0.0 & 1.3 & 1.3 & 1.3 & 4.0 & 14.1$_{\pm 0.3}$ & 10.7$_{\pm 0.2}$ & 12.6$_{\pm 0.2}$ \\
Human-Video & R3M & 48.0 & 0.0 & 0.0 & 0.0 & 0.0 & 0.0 & 8.0 & 2.7 & 2.7 & 1.3 & 12.3$_{\pm 1.4}$ & 6.3$_{\pm 0.9}$ & 9.7$_{\pm 0.6}$ \\
& D4R-Align & 89.3 & 1.3 & 0.0 & 0.0 & 0.0 & 0.0 & 8.0 & 6.7 & 0.0 & 1.3 & 14.5$_{\pm 1.0}$ & 10.7$_{\pm 0.2}$ & 12.8$_{\pm 0.6}$ \\
& R3M-Align & 90.7 & 0.0 & 1.3 & 0.0 & 0.0 & 0.0 & 2.7 & 13.3 & 4.0 & 0.0 & 12.9$_{\pm 0.7}$ & 11.2$_{\pm 0.7}$ & 12.2$_{\pm 0.3}$ \\
\midrule
& OpenVLA & 77.3 & 0.0 & 0.0 & 0.0 & 0.0 & 0.0 & 6.7 & 5.3 & 2.7 & 0.0 & 16.9$_{\pm 1.3}$ & 9.2$_{\pm 0.7}$ & 13.6$_{\pm 0.8}$ \\
& RDT & 100.0 & 29.3 & 4.0 & 0.0 & 0.0 & 0.0 & 8.0 & 2.7 & 0.0 & 0.0 & 10.4$_{\pm 0.5}$ & 14.4$_{\pm 0.9}$ & 12.1$_{\pm 0.4}$ \\
& $\pi_0$ & 97.3 & 0.0 & 1.3 & 0.0 & 0.0 & 0.0 & 14.7 & 5.3 & 1.3 & 0.0 & *21.7$_{\pm 0.4}$ & 12.0$_{\pm 0.9}$ & *17.5$_{\pm 0.4}$ \\
Foundation & LLARVA & 24.0 & 0.0 & 0.0 & 0.0 & 0.0 & 0.0 & 0.0 & 0.0 & 0.0 & 0.0 & 4.5$_{\pm 0.1}$ & 2.4$_{\pm 0.0}$ & 3.6$_{\pm 0.1}$ \\
VLA & 3D-LOTUS & 85.3 & 0.0 & 1.3 & 0.0 & 0.0 & 0.0 & 0.0 & 4.0 & 0.0 & 0.0 & 3.2$_{\pm 0.5}$ & 9.1$_{\pm 0.7}$ & 6.2$_{\pm 0.5}$ \\
& 3D-LOTUS++ & 90.7 & 30.7 & 0.0 & 0.0 & 5.3 & 1.3 & 8.0 & 8.7 & 6.7 & 0.0 & 13.6$_{\pm 1.0}$ & *15.1$_{\pm 1.1}$ & 14.4$_{\pm 1.0}$ \\
& SAM2Act & 92.0 & 49.3 & 0.0 & 0.0 & 0.0 & 0.0 & 1.3 & 6.7 & 4.0 & 5.3 & 13.1$_{\pm 0.4}$ & 15.9$_{\pm 1.3}$ & 14.0$_{\pm 0.7}$ \\
& VoxPoser & 32.0 & 76.0 & 0.0 & 8.0 & 0.0 & 0.0 & 0.0 & 1.3 & 0.0 & 4.0 & 18.1$_{\pm 0.4}$ & 12.1$_{\pm 0.4}$ & 15.6$_{\pm 0.2}$ \\
\midrule
ICL-based & X-ICM & 98.7 & 20.0 & 6.7 & 9.3 & 0.0 & 6.7 & 16.0 & 2.7 & 5.3 & 4.0 & 28.6$_{\pm 1.9}$ & 16.9$_{\pm 1.3}$ & 23.5$_{\pm 1.6}$ \\
\midrule
  & \textbf{Ours}  & 96.0 & 20.3 & 6.7 & 10.0 & 0.0 & 16.0 & 24.0 & 8.0 & 0.0 & 4.0 & \textbf{32.5$_{\pm 1.5}$} & \textbf{18.5$_{\pm 1.0}$} & \textbf{26.4$_{\pm 1.3}$} \\
\bottomrule
\end{tabular}
}
\end{small}
\end{center}
\vspace{-0.4cm}

\end{table*}

\subsection{Skill-Augmented In-Context Learning}

Retrieved demonstrations are formatted with skill--action alignment:
\begin{equation}
[\texttt{instr}_i, \texttt{obs}_i, \texttt{plan}_i] \rightarrow [(s_1, a_1), \ldots, (s_K, a_K)]
\end{equation}

The skill labels serve as interpretable intermediate representations explaining what each action accomplishes and how it connects to the task goal. For the query task, we present instruction, observation, and the suggested plan from the planning agent.

Crucially, while labels guide understanding of action semantics and execution order, the LLM outputs only executable actions without labels. This ensures: (1) direct compatibility with the robot controller; (2) implicit skill-level reasoning even when exact decomposition is uncertain; (3) generalization to novel skill combinations.

\begin{figure*}[!ht]
  \begin{center}
    \centerline{\includegraphics[width=\textwidth]{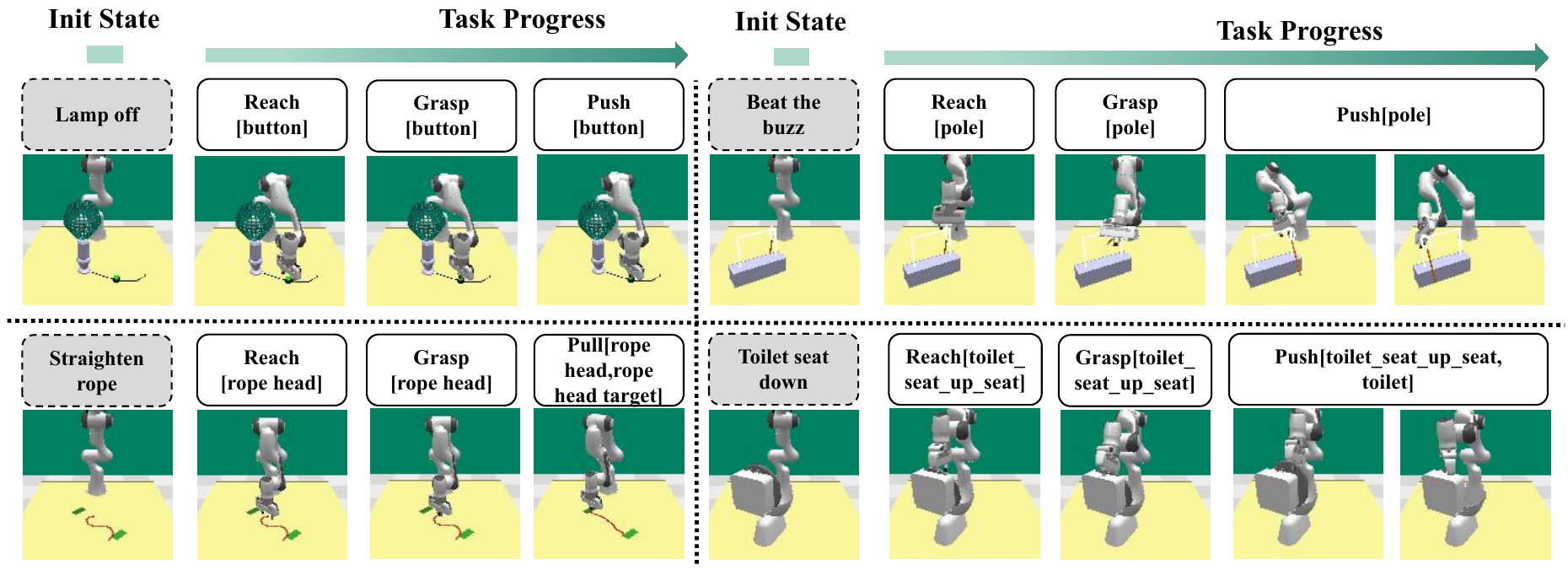}}
    \caption{
Visualization of our compositional skill reasoning process on four representative unseen tasks from the AGNOSTOS benchmark.
    }
    \label{sim}
  \end{center}
    \vspace{-0.6cm}
\end{figure*}

The system prompt emphasizes that demonstrations include skill annotations for reasoning guidance, but the model should output only action sequences. Through this formulation, our approach elevates cross-task transfer from trajectory similarity to composable skill structures, explicitly activating reasoning about skill composition rather than superficial pattern matching.

\section{Experiments}

\subsection{Implementation Details}

We evaluate our approach for zero-shot cross-task robotic manipulation generalization across two settings: the AGNOSTOS benchmark, which is the first simulation testbed specifically designed for cross-task zero-shot manipulation evaluation, and a real-world experimental setup using a UFACTORY xArm6 robotic arm. The hyperparameters $\lambda$, $\alpha$, $\beta$, and $\gamma$ in Dual-Library Demonstration Retrieval are 0.7, 0.5, 0.5, and 0.03, respectively. We use 20 in-context demonstrations in total, with $k_{\text{sim}}=17$ dynamically retrieved demonstrations and $k_{\text{cov}}=3$ coverage-aware demonstrations. We primarily employ Qwen2.5-VL as the planning agent in offline Atomic Skill Collection, and we use the Qwen-2.5-7B-Instruct model as the planning agent and LLM in the reasoning phase. These models are deployed on two A6000 GPUs.

\subsection{Main Results}
To systematically assess the performance of our method, we conduct experiments on the AGNOSTOS benchmark~\cite{zhou2025exploring}, which comprises 23 unseen tasks organized into two difficulty tiers: Level-1 and Level-2.

\paragraph{Baseline Methods.} We compare against diverse Vision-Language-Action (VLA) approaches spanning four distinct categories:

\textbf{(1) Foundation VLA Models.} This category encompasses models that leverage either large-scale cross-embodiment robotic datasets or are architecturally grounded in pretrained LLMs/VLMs. These approaches aim to achieve broad manipulation capabilities through scale and language-vision grounding. Representative methods include OpenVLA~\cite{kim2024openvla}, which pioneers open-source VLA development; RDT~\cite{liu2024rdt}, employing robot-specific transformers; $\pi_0$~\cite{black2024pi_0}, combining VLM-based flow matching with multi-platform training; LLARVA~\cite{niu2024llarva}, utilizing instruction tuning on cross-embodiment data; SAM2Act~\cite{fang2025sam2act} and 3D-LOTUS++~\cite{garcia2024towards}, leveraging advanced visual representations; and VoxPoser~\cite{huang2023voxposer}, synthesizing composable value maps via MLLMs.

\textbf{(2) Human-Video Pretrained VLA Models.} These methods exploit large-scale human activity video datasets~\cite{grauman2022ego4d,goyal2017something} to learn rich object-interaction priors, subsequently adapting these representations for robotic control through downstream finetuning. We evaluate R3M~\cite{nair2022r3m} and D4R~\cite{zhou2024mitigating}, along with their RLBench-adapted variants R3M-Align~\cite{zhou2024mitigating} and D4R-Align~\cite{zhou2024mitigating}.

\textbf{(3) In-Domain Trained VLA Models.} This category includes methods trained from scratch on RLBench's 18 standard training tasks using task-specialized architectures. These approaches serve as strong upper bounds since they operate without cross-domain distribution shift. We evaluate PerAct~\cite{shridhar2023perceiver}, RVT~\cite{goyal2023rvt}, RVT2~\cite{goyal2024rvt}, Sigma-Agent~\cite{chen2024sigmaagent}, and Instant Policy~\cite{vosylius2024instant}---the latter being a within-task in-context learning method based on graph diffusion over structured demonstrations.

\textbf{(4) In-Context Learning Approaches for Cross-Task Manipulation.} We include X-ICM~\cite{zhou2025exploring} as a representative method that constructs in-context prompts from seen task demonstrations to drive LLM-based action prediction for unseen tasks in a zero-shot manner.

As shown in \cref{tab:main_results}, we conduct comprehensive evaluation across all 23 unseen tasks from the AGNOSTOS benchmark to assess both accuracy and generalization capability. For fair comparison, we follow the standard AGNOSTOS evaluation protocol, running 25 rollouts for each of 3 different seeds per task, resulting in 75 episodes per task. Our method achieves the highest overall success rates across both difficulty levels. Notably, we attain success rates exceeding 60\% on four distinct tasks, including Microwave, Seat, LampOff, and USB, whereas other individual baselines achieve this threshold on at most three tasks. This demonstrates that our approach of constructing atomic skill-action alignment pairs, which preserves executable low-level actions while providing compositional and reasoned intermediate representations, effectively stimulates the model's reasoning capabilities over skill composition and execution ordering, thereby enhancing zero-shot cross-task generalization. Figure~\ref{sim} visualizes our method's reasoning process on four unseen tasks: \texttt{lamp\_off}, \texttt{straighten\_rope}, \texttt{beat\_the\_buzz}, and \texttt{toilet\_seat\_down}, providing intuitive illustration of our compositional skill reasoning for cross-task manipulation generalization.

\subsection{Ablation Analysis}

In \cref{table:ablation}, we progressively incorporate each core component to demonstrate the effectiveness of our framework. 
\emph{Baseline (BS.):} The first row presents the baseline that randomly selects in-context demonstrations from seen tasks without any retrieval strategy, achieving 21.6\% overall success rate. 

\emph{Dynamic Demonstrations Library (DDL.):} Adding task-adaptive retrieval based on visual-semantic and skill-sequence matching improves performance to 23.3\%, validating that relevant demonstration selection enhances transferable action patterns. 

\emph{Coverage-aware Demonstrations Library (CDL.):} Further incorporating the static library that supplements missing skill patterns raises performance to 24.9\%, demonstrating the importance of skill coverage completeness. 

\emph{Skill-Augmented In-Context Learning (SA-ICL):} Finally, explicit skill--action alignment annotations yield our best performance of 26.4\%, confirming that structured skill representations effectively elicit compositional reasoning beyond trajectory imitation.

\begin{table}[ht]
\caption{Ablation study on core components. BS is Baseline, DDL denotes the Dynamic Demonstrations Library; CDL denotes the Coverage-aware Demonstrations Library; SA-ICL denotes Skill-Augmented In-Context Learning.}
\label{table:ablation}
\begin{center}
\begin{small}
\begin{tabular}{ccccccc}
\toprule
\textbf{BS.} & \textbf{DDL.} & \textbf{CDL.} & \textbf{SA-ICL} & \textbf{L-1} & \textbf{L-2} & \textbf{All} \\
\midrule
\checkmark &  &  &  & 26.9 & 14.8 & 21.6 \\
\checkmark & \checkmark &  &  & 29.5 & 17.2 & 23.3 \\
\checkmark & \checkmark &\checkmark & & 30.2 & 18.1 & 24.9 \\
\checkmark & \checkmark & \checkmark &\checkmark & \textbf{32.5} & \textbf{18.5} & \textbf{26.4} \\
\bottomrule
\end{tabular}
\end{small}
\end{center}
\vspace{-0.4cm}
\end{table}

\begin{figure*}[!t]
  \begin{center}
    \centerline{\includegraphics[width=\textwidth]{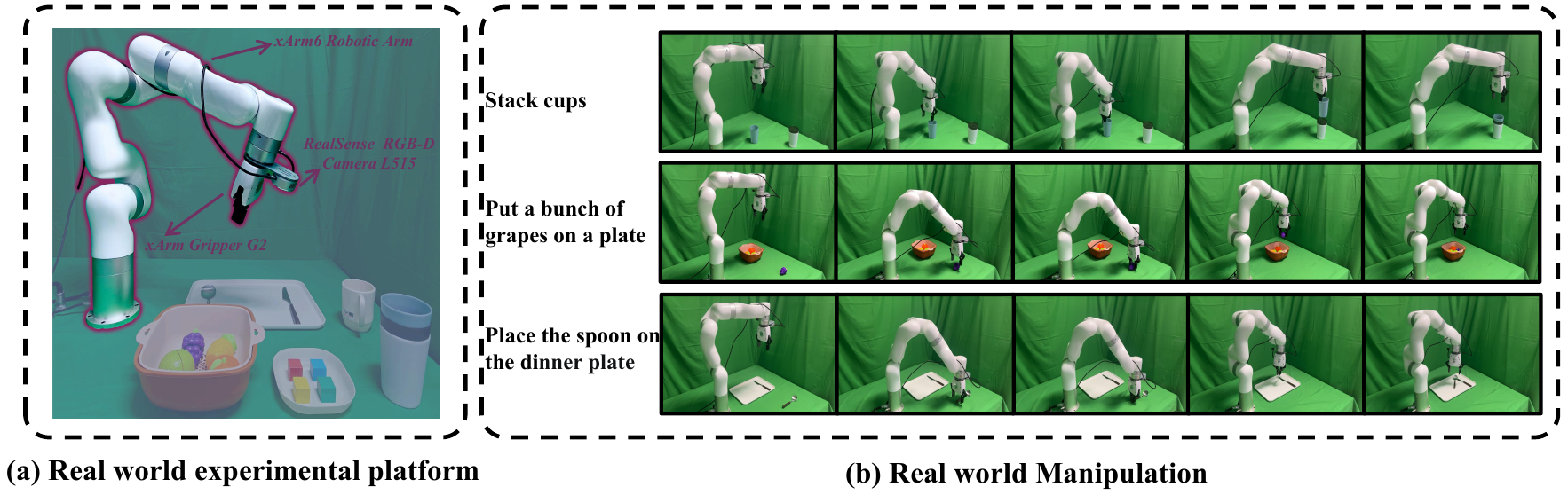}}
    \caption{
   Visualization of our real world experiments.(a) Our real world experimental platform, consisting of a 6-DoF xArm6 arm equipped with a gripper and a RGB-D camera. (b) The results on real-world manipulation tasks.
    }
    \label{realmain}
  \end{center}
  \vspace{-0.6cm}
\end{figure*}

Furthermore, we analyze the impact of visual encoders and LLM backbones. \cref{table:visual_llm} (upper portion) compares performance when replacing DINOv3~\cite{simeoni2025dinov3} with alternative general-purpose visual models such as CLIP~\cite{radford2021learning} and DINOv2~\cite{oquab2023dinov2}, demonstrating that static scene semantic extraction capability influences our method's zero-shot cross-task generalization, underscoring the importance of scene understanding. The lower portion examines different LLM backbones: Llama3.0-8B-Instruct~\cite{grattafiori2024llama}, Ministral-8B-Instruct~\cite{jiang2023mistral}, and InternLM3-8B-Instruct~\cite{cai2024internlm2}. Results reveal that backbone capability significantly impacts generalization, emphasizing that cross-task generalization improves with enhanced LLM reasoning ability. Additional ablation details are provided in Appendix \ref{ablationsupp}.

\begin{table}[t]
\caption{Impact of visual encoders (upper) and LLM backbones (lower) on cross-task generalization performance.}
\label{table:visual_llm}

\begin{center}
\begin{small}
\begin{tabular}{lccc}
\toprule
\textbf{Component} & \textbf{L-1} & \textbf{L-2} & \textbf{Overall} \\
\midrule
\multicolumn{4}{l}{\textit{Visual Encoder}} \\
CLIP & 30.5 & 17.4 & 24.8 \\
DINOv2 & 31.7 & 17.9 & 25.7 \\
DINOv3 & \textbf{32.5} & \textbf{18.5} & \textbf{26.4} \\
\midrule
\multicolumn{4}{l}{\textit{LLM Backbone}} \\
Llama3.0-8B & 28.4 & 14.6 & 22.4 \\
Ministral-8B & 24.3 & 14.8 & 20.2 \\
InternLM3-8B & 30.1 & 16.3 & 24.1 \\
Qwen2.5-7B & \textbf{32.5} & \textbf{18.5} & \textbf{26.4} \\
\bottomrule
\end{tabular}
\end{small}
\end{center}
\vspace{-0.6cm}
\end{table}

\subsection{Real-World Experiments}

To validate real-world efficacy, we conduct real-world manipulation experiments. As shown in Figure~\ref{realmain}(a), we use a UFACTORY xArm6 arm equipped with an UFACTORY xArm Gripper G2, with RGB images and depth maps captured via a RealSense L515 camera. We provide additional details in Appendix \ref{realsupp}.

\begin{table}[t]
\caption{Real-world zero-shot cross-task manipulation results. Success rates (\%) over 25 trials per task.}
\label{table:realworld}
\begin{center}
\begin{small}
\begin{tabular}{lc}
\toprule
\textbf{Task} & \textbf{Success Rate (\%)} \\
\midrule
Stack cups & 30 \\
Put a bunch of grapes on a plate & 70 \\
Place the spoon on the dinner plate & 20 \\
Stack blocks & 20 \\
Throw the garbage into the trash can  & 40  \\
\midrule
\textbf{Average} & \textbf{36} \\
\bottomrule
\end{tabular}
\end{small}
\end{center}
\vspace{-0.8cm}
\end{table}

 We evaluate five physical manipulation tasks: stacking cups, putting a bunch of grapes on a plate, placing the spoon on the dinner plate, stacking blocks, and throwing the garbage into the trash can. Results shown in \cref{table:realworld} demonstrate our method's  zero-shot cross-task performance in real-world settings. Figure~\ref{realmain}(b) illustrates a subset of successful cases, with additional visualization results that include both successful and failed cases can be found in Appendix \ref{realsupp} and videos are provided in the supplementary materials.

\section{Conclusion}

We present \emph{Decompose and Recompose}, a skill reasoning framework for zero-shot cross-task robotic manipulation. Our core contribution is introducing atomic skill--action pairs as intermediate representations, bridging high-level task semantics and low-level controls. Through a dual-library retrieval strategy combining task-adaptive dynamic retrieval with coverage-aware static complementation, we construct skill-comprehensive demonstration sets that explicitly elicit the LLM's compositional reasoning capabilities. Experiments on the AGNOSTOS benchmark and real-world environments validate our method's effectiveness for cross-task generalization.

\section*{Impact Statement}
This paper presents work that aims to advance the field
of Cross-Task Robotic Manipulation Generalization. There are many potential societal consequences of our work, none of which we feel must be specifically highlighted here.

\bibliography{example_paper}
\bibliographystyle{icml2026}

\newpage
\appendix
\onecolumn

\section{Methodology and Implementation Details}
\label{canshudingyi}

\subsection{Low-level control interface}
We use a standard RLBench control interface that executes an end-effector target pose via motion planning and applies a discrete gripper open/close command. Concretely, each continuous control step is represented as
\begin{equation}
\mathbf{u} = [\mathbf{p}, \mathbf{q}, g] \in \mathbb{R}^3 \times \mathbb{H} \times \{0,1\},
\end{equation}
where $\mathbf{p}\in\mathbb{R}^3$ is the end-effector position in the robot base frame, $\mathbf{q}\in\mathbb{H}$ is a unit quaternion specifying orientation, and $g$ is a binary gripper command (e.g., $g{=}1$ for open, $g{=}0$ for close).

\subsection{Discrete action format for the LLM}
For LLM interaction, we discretize translation and rotation into integer bins and represent each action as a 7-tuple:
\begin{equation}
\mathbf{a} = [i_x,i_y,i_z,i_r,i_p,i_\psi,g],
\end{equation}
where $(i_x,i_y,i_z)$ are voxel indices for translation, $(i_r,i_p,i_\psi)$ are discrete bins for Euler angles (roll, pitch, yaw), and $g$ is the gripper command.

\subsection{Encoding: continuous control $\rightarrow$ discrete LLM tokens}
\paragraph{Translation discretization.}
Let the workspace bounds be an axis-aligned box
\begin{equation}
\mathbf{b}_{\min} \in \mathbb{R}^3,\qquad \mathbf{b}_{\max} \in \mathbb{R}^3,
\end{equation}
and let $V$ be the number of uniform bins per axis (we use $V{=}100$). Define per-axis resolution
\begin{equation}
\mathbf{r} = (\mathbf{b}_{\max}-\mathbf{b}_{\min})/V.
\end{equation}
Given a continuous position $\mathbf{p}$, the voxel index is computed as
\begin{equation}
\mathbf{i} = \big\lfloor (\mathbf{p}-\mathbf{b}_{\min}) \oslash \mathbf{r} \big\rfloor,
\end{equation}
where $\oslash$ denotes elementwise division and the result is clipped to $[0, V{-}1]$ per axis.

\paragraph{Rotation discretization.}
We convert the quaternion $\mathbf{q}$ into Euler angles (degrees) $\boldsymbol{\theta}=[\theta_r,\theta_p,\theta_y]$ in a fixed convention. We then quantize angles with resolution $\Delta$ (we use $\Delta{=}5^\circ$):
\begin{equation}
\mathbf{k} = \big\lfloor (\boldsymbol{\theta} + 180^\circ)/\Delta \big\rfloor,
\end{equation}
and clip each element to $[0, \lfloor 360^\circ/\Delta \rfloor - 1]$.

\paragraph{Gripper.}
The gripper command is kept discrete as $g \in \{0,1\}$.

\subsection{Decoding: discrete LLM tokens $\rightarrow$ continuous control}
Given an LLM-predicted discrete action $\mathbf{a}=[\mathbf{i},\mathbf{k},g]$, we recover a continuous control target as follows.

\paragraph{Translation reconstruction.}
We map voxel indices to the center of the corresponding voxel cell:
\begin{equation}
\mathbf{p} = \mathbf{b}_{\min} + \mathbf{r}\odot \mathbf{i} + \mathbf{r}/2,
\end{equation}
where $\odot$ denotes elementwise multiplication.

\paragraph{Rotation reconstruction.}
We recover Euler angles from bins by
\begin{equation}
\boldsymbol{\theta} = \Delta \cdot \mathbf{k} - 180^\circ,
\end{equation}
then convert $\boldsymbol{\theta}$ back to a quaternion $\mathbf{q}$ using the same Euler convention as in the encoding step.

\paragraph{Execution.}
Finally, the low-level control command is assembled as $\mathbf{u}=[\mathbf{p},\mathbf{q},g]$ and executed by the RLBench action mode (end-effector pose via planning + discrete gripper). This completes the closed loop between the LLM interface and the robot controller.

\subsection{Practical notes}
In practice, clipping ensures indices remain valid even when the predicted pose slightly exceeds workspace limits. Using voxel centers provides a consistent inverse mapping and reduces discretization bias. Rotation discretization is similarly bounded to guarantee valid bins, while quaternion normalization is applied after reconstruction to ensure a valid rotation.

\subsection{Algorithm details}
\begin{algorithm}[tb]
   \caption{Decompose and Recompose: Skill-Based Cross-Task Manipulation}
   \label{alg:dual_library}
\begin{algorithmic}[1]
   \REQUIRE Query observation $o^q$, instruction $\ell^q$, demonstration library $\mathcal{D}_{\text{seen}}$, Coverage-aware Demonstrations Library $\mathcal{L}_{\text{cov}}$
   \ENSURE Predicted action sequence $\{a^q_1, \ldots, a^q_T\}$
   
   \STATE \textbf{// Phase 1: Plan Generation}
   \STATE Extract scene state from $o^q$; generate skill sequence $\hat{S} \leftarrow \pi_{\text{plan}}(\ell^q, \text{scene state})$
   
   \STATE \textbf{// Phase 2: Dynamic Library Retrieval}
   \STATE Encode query image: $f^q \leftarrow \phi_{\text{vis}}(o^q) / \|\phi_{\text{vis}}(o^q)\|_2$
   \FOR{each demonstration $d_i \in \mathcal{D}_{\text{seen}}$}
       \STATE Compute visual similarity $s^{\text{vis}}_i$ and plan similarity $s^{\text{plan}}_i$
       \STATE Fuse scores: $s_i \leftarrow \alpha \cdot \tilde{s}^{\text{vis}}_i + (1-\alpha) \cdot s^{\text{plan}}_i$
   \ENDFOR
   \STATE Select top-$k_{\text{sim}}$ demonstrations: $\mathcal{D}_{\text{dyn}}$
   
   \STATE \textbf{// Phase 3: Static Library Coverage Completion}
   \STATE Compute covered tokens $\mathcal{C}$ and missing patterns $\mathcal{G} \leftarrow \mathcal{T}(\hat{S}) \setminus \mathcal{C}$
   \WHILE{$|\mathcal{D}_{\text{cov}}| < k_{\text{cov}}$ \AND $\mathcal{G} \neq \emptyset$}
       \STATE Select $d^* \leftarrow \arg\max_d \text{Score}(d)$ from $\mathcal{L}_{\text{cov}}$; update $\mathcal{D}_{\text{cov}}$, $\mathcal{C}$, $\mathcal{G}$
   \ENDWHILE
   
   \STATE \textbf{// Phase 4: Skill-Augmented In-Context Learning}
   \STATE Merge demonstrations: $\mathcal{D} \leftarrow \mathcal{D}_{\text{dyn}} \cup \mathcal{D}_{\text{cov}}$
   \STATE \textbf{Return} $\{a^q_1, \ldots, a^q_T\} \leftarrow \text{LLM}(\ell^q, o^q, \hat{S}, \mathcal{D})$
\end{algorithmic}

\end{algorithm}
Algorithm~\ref{alg:dual_library} summarizes our complete pipeline. The algorithm proceeds in four phases: (1) \textbf{Plan Generation} produces a predicted skill sequence $\hat{S}$ that guides subsequent retrieval; (2) \textbf{Dynamic Demonstrations Library} selects task-adaptive demonstrations by fusing visual similarity with plan-based similarity; (3) \textbf{Coverage-Aware Completion} greedily supplements missing skill patterns from the static library using IDF-weighted scoring with length regularization; (4) \textbf{Skill-Augmented ICL} constructs the final prompt where skill labels provide compositional reasoning guidance while the model outputs only executable actions.

The dual-library design achieves complementary objectives: the dynamic library provides \emph{task-adaptive} demonstrations based on visual and semantic relevance, while the static library ensures \emph{skill-coverage sufficiency}. This combination enables the LLM to reason over composable skill structures rather than merely imitating trajectory patterns.

\begin{table}[ht]
\caption{Ablation study on the number of demonstrations (upper) and coverage-aware demonstrations (lower) in in-context learning.}
\label{table:demos_num}

\begin{center}
\begin{small}
\begin{tabular}{lccc}
\toprule
\textbf{Component} & \textbf{L-1} & \textbf{L-2} & \textbf{Overall} \\
\midrule
\multicolumn{4}{l}{\textit{Demos per ICL}} \\
0  & 0.0  & 0.0  & 0.0 \\
5  & 27.3 & 13.1 & 19.8 \\
10 & 30.0 & 15.5 & 23.7 \\
15 & 31.2 & 16.8 & 24.9 \\
20 & \textbf{32.5} & \textbf{18.5} & \textbf{26.4} \\
\midrule
\multicolumn{4}{l}{\textit{Coverage-aware Demos per ICL}} \\
0 & 31.0  & 17.1  & 24.9 \\
1 & 31.6 & 17.7 & 25.5 \\
2 & 32.1 & 18.2 & 26.0 \\
3 & \textbf{32.5} & \textbf{18.5} & \textbf{26.4} \\
4 & 32.3 & 18.1 & 26.1 \\

\bottomrule
\end{tabular}
\end{small}
\end{center}
\vspace{-0.4cm}
\end{table}

\begin{figure*}[!ht]
  \begin{center}
    \centerline{\includegraphics[width=\textwidth]{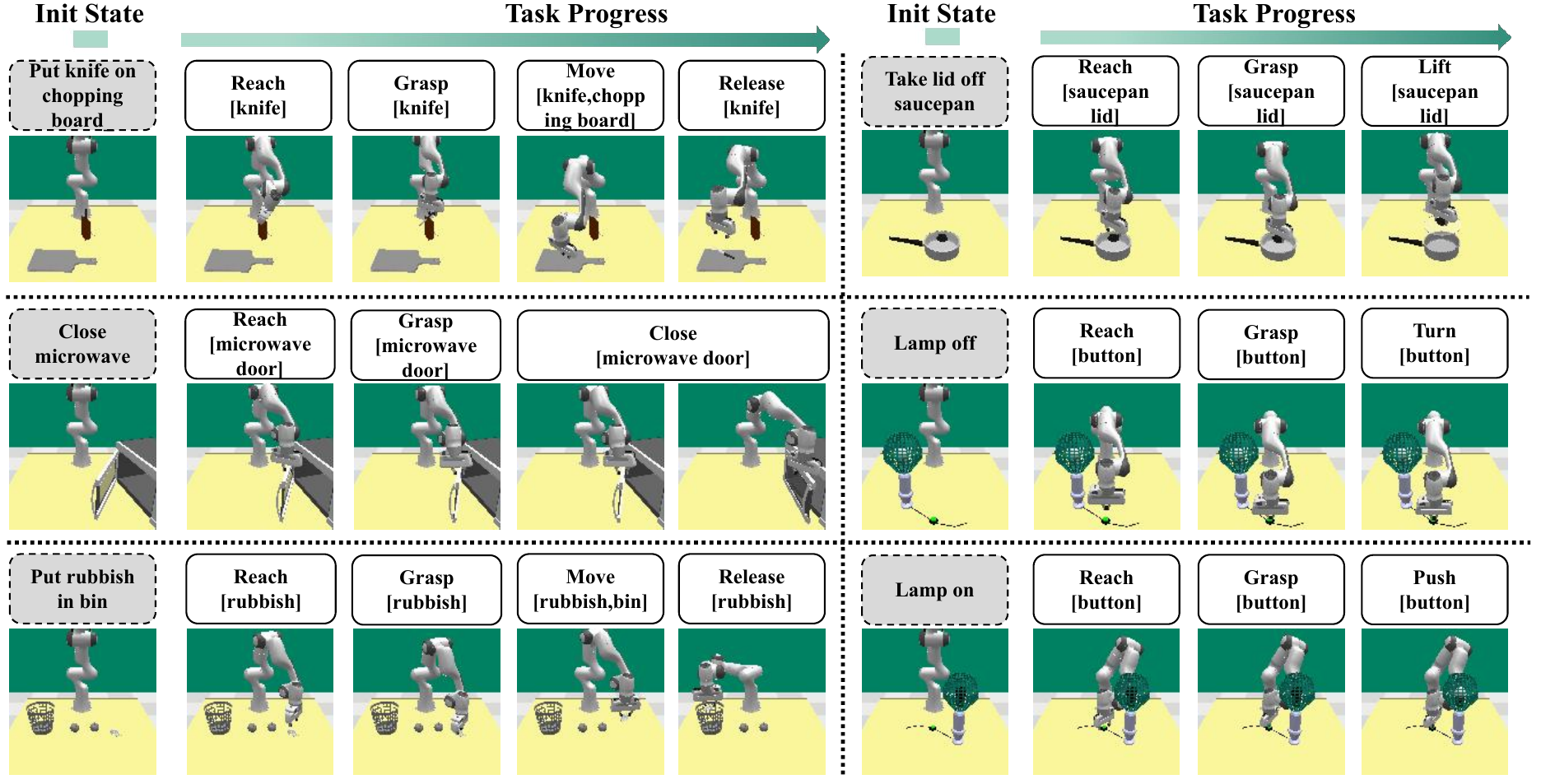}}
    \caption{
Visualization of our compositional skill reasoning process on some representative unseen tasks from the AGNOSTOS benchmark.
    }
    \label{sim_supp}
  \end{center}
    \vspace{-0.4cm}
\end{figure*}

\subsection{More visual results}
Figure~\ref{sim_supp} illustrates the reasoning process of our method on six unseen task of the AGNOSTOS benchmark.

\section{More Ablation}
\label{ablationsupp}
We conduct additional ablation studies to analyze the impact of demonstration quantity on cross-task generalization performance. Results are summarized in \cref{table:demos_num}.

\paragraph{Number of in-context demonstrations.}
The upper portion of \cref{table:demos_num} examines how the total number of demonstrations in the in-context prompt affects performance. When no demonstrations are provided, the model completely fails to generate valid action sequences, achieving 0\% success rate across all tasks. This confirms that in-context demonstrations are essential for cross-task manipulation, as the LLM cannot infer appropriate action patterns from task instructions alone.
As the number of demonstrations increases from 5 to 20, we observe consistent improvements: overall success rate rises from 19.8\% to 26.4\%, with Level-1 tasks improving from 27.3\% to 32.5\% and Level-2 tasks from 13.1\% to 18.5\%. This trend indicates that richer demonstration sets provide more comprehensive skill--action patterns for the model to reason over.

\begin{figure*}[!t]
  \begin{center}
    \centerline{\includegraphics[width=\textwidth]{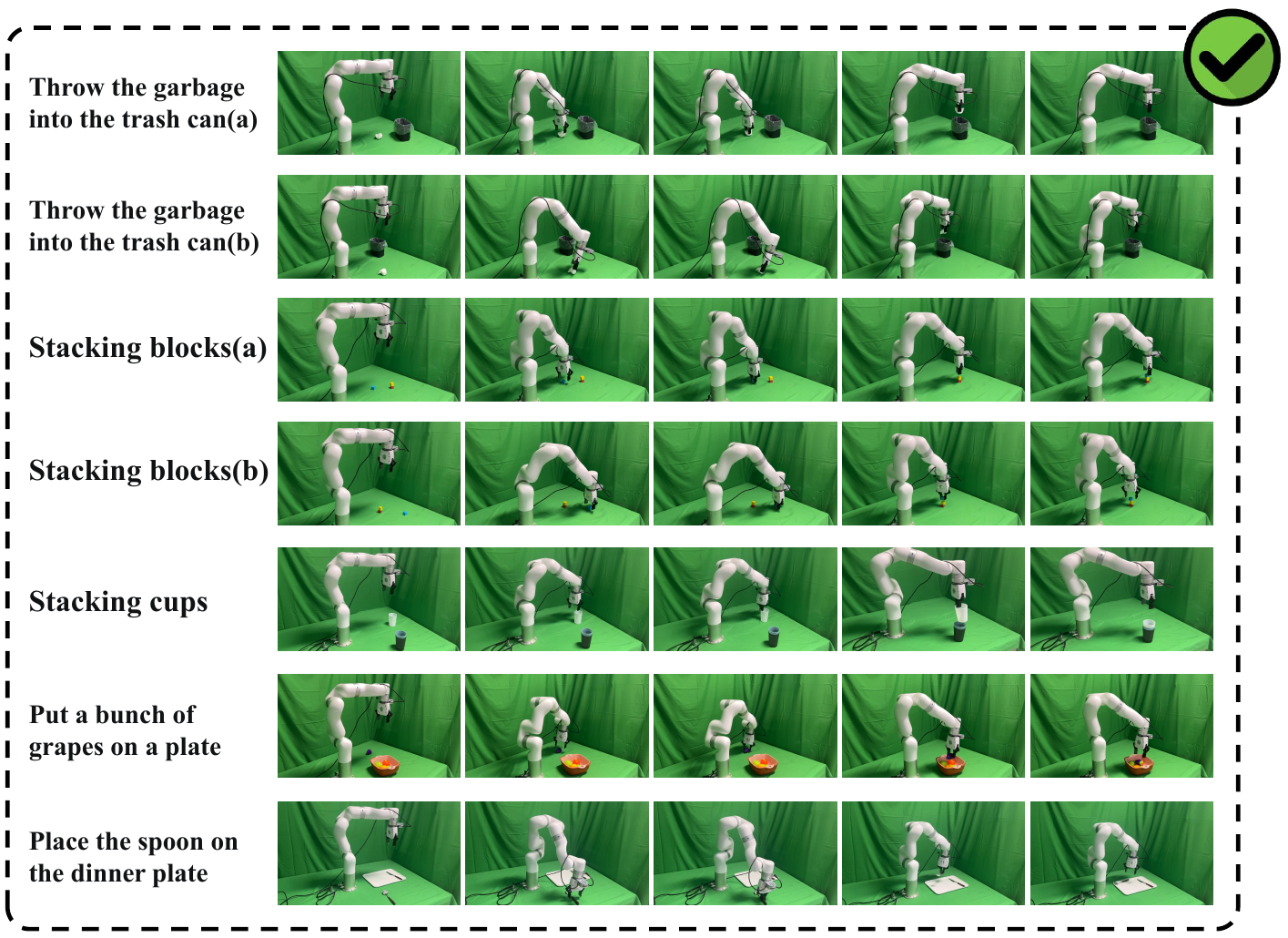}}
    \caption{
   Successful cases of our real world experiments.
    }
    \label{real_suppfig}
  \end{center}
  \vspace{-0.4cm}
\end{figure*}

\paragraph{Number of coverage-aware demonstrations.}
The lower portion of \cref{table:demos_num} investigates the contribution of coverage-aware demonstrations, which supplement missing skill patterns not captured by the dynamic retrieval. Starting from a baseline of 0 coverage-aware demonstrations, i.e., relying solely on dynamically retrieved examples, performance stands at 24.9\%. 

Adding coverage-aware demonstrations yields steady improvements: 1 demo achieves 25.5\%, 2 demonstrations reach 26.0\%, and 3 demonstrations attain the best overall performance of 26.4\%. Notably, performance slightly decreases with 4 demonstrations, suggesting that excessive static demonstrations may introduce noise or dilute the relevance of task-adaptive examples. This indicates an optimal balance exists between ensuring skill coverage completeness and maintaining demonstration relevance.

These results validate two key design choices: (1) sufficient demonstration quantity is necessary to expose diverse skill--action patterns, and (2) the coverage-aware static library effectively complements dynamic retrieval by filling critical skill gaps, but should be used judiciously to avoid over-saturation.

\section{Real-world experiments}
\label{realsupp}
\textbf{Experimental setup:} we use a UFACTORY xArm6 arm equipped with an UFACTORY xArm Gripper G2, with RGB images and depth maps captured via a RealSense L515 camera. We evaluate five physical manipulation tasks: stacking cups, putting a bunch of grapes on a plate, placing the spoon on the dinner plate, stacking blocks, and throwing the garbage into the trash can. We collect 20 demonstrations per task, and during testing, construct cross-task in-context prompts using demonstrations from the remaining four tasks, enabling zero-shot generalization. We employ Qwen2.5-7B-Instruct as the backbone with 20 seen demonstrations. Each task is executed 25 times, with average success rates reported.

\begin{figure*}[!ht]
  \begin{center}
    \centerline{\includegraphics[width=\textwidth]{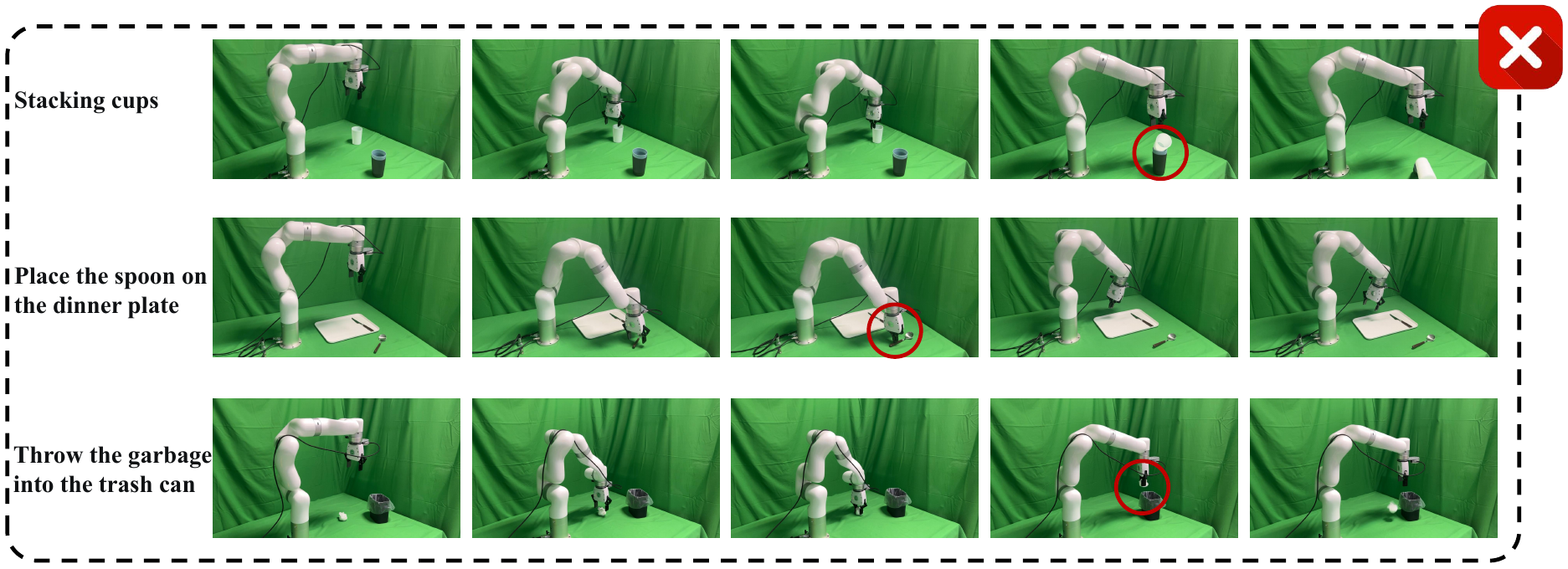}}
    \caption{
   Failing cases of our real world experiments.
    }
    \label{real_suppfig_fail}
  \end{center}
  \vspace{-0.4cm}
\end{figure*}

 Figure~\ref{real_suppfig} shows successful examples from our real-world task experiments. Across all five tasks, our method demonstrates the ability to generate appropriate skill sequences and execute precise actions in diverse manipulation scenarios. For instance, in the \textit{stack cups} and \textit{stack blocks} tasks, the robot accurately grasps objects and places them at the correct positions. These results indicate that our framework effectively transfers skills learned from seen tasks to novel manipulation scenarios in real-world environments.
 
 Additionally, as illustrated in Figure~\ref{real_suppfig_fail}, we present several representative failure cases from our real-world experiments. In the first row, during the \textit{stack cups} task, the robot arm successfully navigates to the stacking location; however, the cup is grasped at a tilted angle, causing the subsequent placement to fail due to misalignment. The second row depicts a failure in the \textit{place the spoon on the dinner plate} task, where the thin handle of the spoon prevents the gripper from achieving a secure grasp. The third row shows the execution of \textit{throw the garbage into the trash can}. In this case, although the robot successfully grasps the garbage and moves it above the trash can, the end-effector hovers over the edge rather than the center of the bin. Consequently, when the gripper releases, the garbage falls outside the container.

These failure cases highlight critical challenges for cross-task robotic manipulation. First, understanding the 6-DoF pose of real-world objects, including their geometric properties and functional affordances, is essential for reliable grasping and placement. Second, accurate reasoning about 3D spatial relationships, such as determining whether the end-effector is positioned above the center versus the periphery of a target region, remains crucial for successful task completion. In future work, we plan to address these limitations by incorporating explicit 3D spatial reasoning and object affordance understanding into our framework.

\end{document}